\pdfoutput=1

\documentclass[11pt]{article}

\usepackage[final]{acl}

\usepackage{times}
\usepackage{latexsym}

\usepackage[T1]{fontenc}

\usepackage[utf8]{inputenc}

\usepackage{microtype}
\usepackage{pifont} 
\newcommand{\treelogo}{\raisebox{5pt}{\includegraphics[scale=0.050]{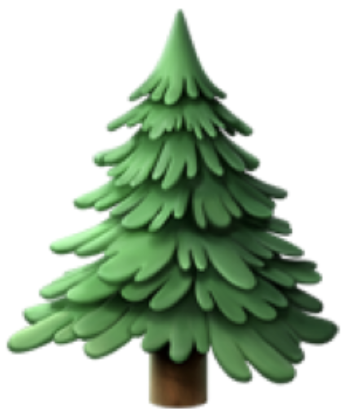}}}
\newcommand{\gtlogo}{\raisebox{3.4pt}{\includegraphics[scale=0.025]{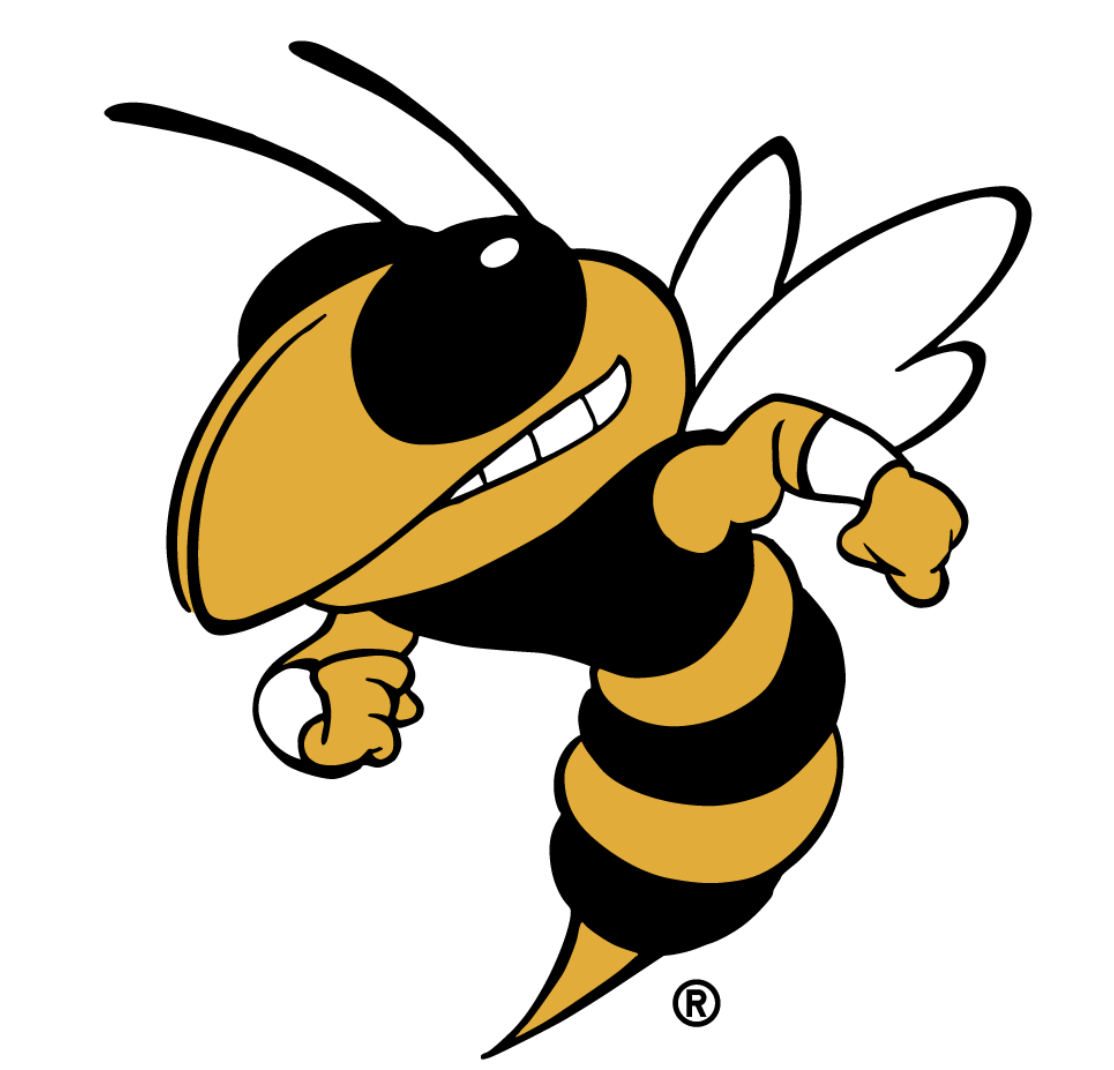}}}
\newcommand{\gt}{\gtlogo}
\newcommand{\stanf}{\treelogo}
\usepackage{soul} 

\usepackage{tipa}
\usepackage{dblfloatfix}
\usepackage{amsmath}
\usepackage{graphicx}
\usepackage{array,booktabs,ragged2e}
\usepackage{multirow}
\usepackage{array}
    \newcolumntype{P}[1]{>{\centering\arraybackslash}p{#1}}
    \newcolumntype{C}[1]{>{\centering\arraybackslash}c{#1}}
    \newcolumntype{L}[1]{>{\centering\arraybackslash}l{#1}}
\usepackage{booktabs}

\newcolumntype{R}[1]{>{\RaggedLeft\arraybackslash}p{#1}}
\newcolumntype{L}[1]{>{\RaggedRight\arraybackslash}p{#1}}
\newcolumntype{M}[1]{>{\centering\arraybackslash}m{#1}}
\usepackage{xcolor}
\definecolor{lpink}{cmyk}{0, 0.7808, 0.4429, 0.1412}
\definecolor{aqua}{cmyk}{0.91, 0, 0.09, 0.36}
\definecolor{ao}{rgb}{0.0, 0.5, 0.0}
\definecolor{amber}{rgb}{1.0, 0.49, 0.0}
\definecolor{dblue}{rgb}{0.0, 0.0, 0.61}
\definecolor{burgundy}{rgb}{0.5, 0.0, 0.13}

\usepackage{inconsolata}

\usepackage{graphicx}
\graphicspath{ {images/} }

\newcommand\SSsize{16,550}
\definecolor{dark_blue}{HTML}{4261B3}

\title{Silent Signals, Loud Impact: LLMs for Word-Sense Disambiguation of Coded Dog Whistles}

\author{
    Julia Kruk\gt \hspace{.5em} Michela Marchini\stanf \hspace{.5em} Rijul Magu\gt \hspace{.5em} Caleb Ziems\stanf \\
    \textbf{David Muchlinski}\gt \hspace{.5em} \textbf{Diyi Yang}\stanf \\   
    \gt Georgia Institute of Technology, \stanf Stanford University \\
    \texttt{\small\{jkruk3, rijul.magu, dmuchlinski3\}@gatech.edu}, \texttt{\small\{marchini, cziems, diyiy\}@stanford.edu}
}

\begin{document}
\maketitle

\begin{abstract}
\vspace*{-1pt}

\textit{\textcolor{red}{Warning: This paper contains content that may be upsetting or offensive to some readers.}}

A dog whistle is a form of coded communication that carries a secondary meaning to specific audiences and is often weaponized for racial and socioeconomic discrimination. Dog whistling historically originated from United States politics, but in recent years has taken root in social media as a means of evading hate speech detection systems and maintaining plausible deniability. In this paper, we present an approach for word-sense disambiguation of dog whistles from standard speech using Large Language Models (LLMs), and leverage this technique to create a dataset of \SSsize{}  high-confidence coded examples of dog whistles used in formal and informal communication. \textbf{Silent Signals}\footnote{\url{huggingface.co/datasets/SALT-NLP/silent_signals}} is the largest dataset of disambiguated dog whistle usage, created for applications in hate speech detection, neology, and political science.

\end{abstract}

\section{Introduction}

\begin{quote}
    {
    ``\textit{Ronald Reagan liked to tell stories of Cadillac-driving 'welfare queens' and 'strapping young bucks' buying T-bone steaks with food stamps. In flogging these tales about the perils of welfare run amok, Reagan always denied any racism and emphasized he never mentioned race.}'' \\\phantom{abc}--- \textbf{Ian Haney-Lopez} (\citeyear{haney2014dog})
    }
\end{quote}

\begin{figure}[t!h]
    \centering
    \includegraphics[width=\columnwidth]{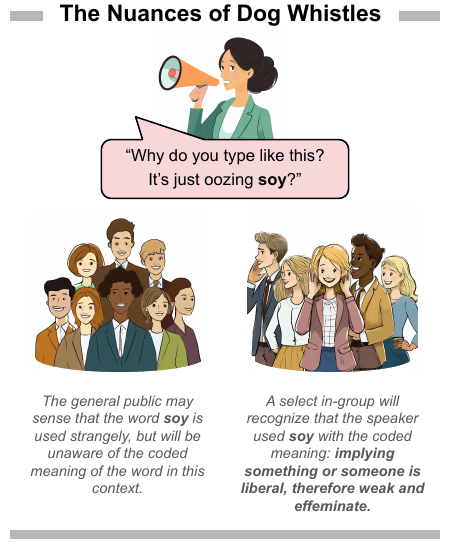}
    \caption{This figure demonstrates the nuances of dog whistle detection as a word can be used in a \textit{coded} or \textit{non-coded} sense. \textit{All illustrations were created using Adobe Firefly.}}
    \label{fig:dog_whistles}
\vspace*{-10pt}
\end{figure}

Dog whistles are coded language which, though seemingly innocuous to the general public, can communicate a covert harmful meaning to a specific in-group \cite{how_dw_work}. Though this coded language appears in all kinds of speech, the idea of the `dog whistle' historically originates in politics \cite{Albertson2014, haney2014dog}. In the United States, political dog whistles gained popularity in the Civil Rights Era following the landmark Brown vs. Board of Education Supreme Court decision, as overt racism became less acceptable and politicians turned to coded language to maintain racial animus in political discussions while maintaining plausible deniability \cite{Saul2018}. Dog whistle use has fluctuated in the last six decades, but their use remains a consistent signal of a speaker's underlying prejudices, whether in the domain of American politics or otherwise. Their use has been shown to successfully provoke the underlying prejudice of target audiences, and wield racial anxiety to steer public opinion and policy \cite{drakulich2020race, wetts2019called}. 

Improved understanding of dog whistles has applications in content moderation, computational social science, and political science. detecting and explaining coded discriminatory speech is a challenging task for NLP systems, as dog whistles famously evade toxicity and hate speech detection \cite{magu2017detecting, magu2018determining, Mendelsohn2023FromDT}. This is because many dog whistle terms have standard vernacular meanings. Consider the example in Figure~\ref{fig:dog_whistles} on the word ``soy," which in most contexts refers to a soybean product, but can also serve as a dog whistle to denigrate liberal or establishment Republican men for perceived feminine attributes, as in ``\textit{That guy has soy face}.'' To study this language, prior work has focused on taxonomically organizing and archiving dog whistles with representative examples \cite{torices2021understanding, Mendelsohn2023FromDT, ryskina2020new, zhu-jurgens-2021-structure}. However, dog whistles can also evolve over time in order to remain covert, a process which has only become more rapid in the age of the internet \cite{dor2004englishization, merchant2001teenagers}.

This work presents a large dataset to track examples of dog whistles in their various forms, and help train language models to do the same. This resource can be used to (1) study how dog whistles emerge and evolve \cite{Saul2018, weimann2020digital}, (2) uncover ways to predict new dog whistle terms from knowledge of old ones, (3) study the prevalence of dog whistles in natural settings, and (4) improve hate speech and toxicity detection systems. As a preliminary step, this work employs LLMs for dog whistle word-sense disambiguation---a new task. We then apply these architectures to create \textbf{Silent Signals}, which is the largest dataset of coded dog whistle examples. It contains \textit{formal} dog whistles from 1900-2023 Congressional records, and \textit{informal} dog whistles from Reddit between 2008-2023. Silent Signals also contains vital contextual information for reliably decoding their hidden meanings and enabling future study of how Dog Whistles are used in discourse. Our contributions include:

\begin{itemize}\itemsep0em
  \item The \textbf{Silent Signals} dataset of \SSsize{} dog whistle examples.
  \item A novel task and verified method for dog whistle word-sense disambiguation.
  \item Experiments with GPT-3.5, GPT-4, Mixtral and Gemini on dog whistle detection. 
  \item The \textbf{Potential Dog Whistle Instance} dataset with over 7 million records from informal and formal communication that contain dog whistle key words, and can be used for further scaling Silent Signals.
\end{itemize}

\section{Related Work}

\paragraph{Hate Speech} 
Prior work categorizes abusive language with two dimensions: target specificity (directed or generalized) and explicitness (explicit or implicit) \cite{waseem2017understanding}. In addition to detecting of explicit language \cite{davidson2017automated, nobata2016abusive}, recent work also labels, detects, and explains the latent meaning behind implicitly abusive language \cite{elsherief2021latent, hartvigsen2022toxigen,breitfeller2019finding,sap2020social,zhou-etal-2023-cobra}, but these works do not primarily focus on dog whistles at scale.

\paragraph{Dog Whistles} Though there is limited prior NLP research on dog whistles, prior work in linguistics has explored the semantics and pragmatics of dog whistles \cite{Saul2018, torices2021understanding, quaranto2022dog, perry2023mating}, and applied agent-based models to the study of the evolution of dog whistle communication \cite{how_dw_work, henderson-mccready-2020-towards}. \citet{Mendelsohn2023FromDT} produced a glossary of over 300 dog whistles used in both the formal and informal settings, and conducted a preliminary survey of the abilities of GPT-3 in the task of dog whistle definition. We extend upon this initial exploration by breaking down \textit{Automatic Dog Whistle Resolution} into sub-tasks of varying complexity, and evaluating LLMs that have been shown to preform well on content moderation tasks \cite{jiang2024mixtral, buscemi2024chatgpt}. The Allen AI Glossary of Dog Whistles \cite{Mendelsohn2023FromDT} is also instrumental in the creation of the Silent Signals dataset presented in this work. Additionally, it is important to note that Dog whistle research in NLP is not limited to American or English-speaking contexts, but extends to coded language in Chinese \citep{xu2021blow} and Swedish \citep{hertzberg-etal-2022-distributional} communication as well.

\paragraph{Political Science Implications}

After the Jim Crow era, once explicitly racist commentary was no longer tolerated  \cite{mendelberg2001race, lasch2016sanctuary}, dog whistles became part of the GOP's ``Southern Strategy'' to maintain racial animus in politics without attracting public ridicule. Although its use dates back to the early 20th century, it is still a very prominent part of American politics \cite{drakulich2020race}. It is a means of political manipulation that encourages people to act on existing biases
and vote for policies against their own interests \cite{wetts2019called, Saul2018}. Prior work has also highlighted that the communication of different messages to different groups makes inferring policy mandates once a candidate assumes office incredibly problematic \cite{goodin2005dog}. To this end, longitudinal dog whistle datasets could facilitate the study of political parties' co-evolution with political, social, and economical events, and improved dog whistle detection could deter ongoing adverse political manipulation.

\paragraph{Word Sense Disambiguation}
Modern Word Sense Disambiguation (WSD) systems can outperform humans \cite{maru2022nibbling, bevilacqua2020breaking, barba2021esc, conia2021framing, kohli2021training}.  WSD tasks are typically treated as multi-label classification problems for resolving the semantic interpretation of target words in context \cite{bevilacqua2021recent, barba2021consec}. A large body of research has focused on designing systems in supervised settings, leveraging pre-trained language models as foundational frameworks \cite{maru2022nibbling, barba2021esc, scarlini2020more, blevins2020moving}. Notably, recent work has explored the use of LLMs for WSD, with findings pointing to strong performance on benchmark evaluations, but still short of levels attained by state-of-the-art models \cite{kocon2023chatgpt}. The detection of pejorative or abusive uses of taboo lexemes has been framed as a WSD task \citep{dinu2021computational,pamungkas2020you,pamungkas2023investigating}, but in these cases, the taboos were overt, and these prior works did not evaluate LLMs. Our study extends the evaluation of LLMs for WSD to contexts where word senses can be deliberately obfuscated or coded. 

\begin{table*}[!th]
    \centering
    \small
    \setlength{\extrarowheight}{2pt}
    \begin{tabular}{||l|p{3in} c c||}
    \hline
        \multicolumn{1}{||c}{\textbf{Dataset}} & \multicolumn{1}{|c}{\textbf{Description}} & \multicolumn{2}{c||}{\textbf{Size}} \\
        ~ & ~ & Informal & Formal \\ \hline\hline
        \textcolor{dark_blue}{\textbf{Potential Instance Dataset}} & Produced via keyword search for dog whistle terms on data collected from Congressional records and Reddit. Used as input data for the creation of \textit{Silent Signals}. & 6,026,910 & 1,088,130 \\
        \textcolor{dark_blue}{\textbf{Synthetic-Detection}} & Manually annotated dataset of dog whistles examples from the \textit{Potential Instance Dataset} used for Dog Whistle Resolution. 50/50 split on positive and negative examples. & 50 & 50\\
        \textcolor{dark_blue}{\textbf{Synthetic-Disambiguation}} & Manually annotated dataset where positive and negative examples are grouped by the dog whistle term they contain. Includes 13 distinct dog whistles. Designed specifically for evaluation on the Dog Whistle Disambiguation task.  & 74 & 50 \\
        \textcolor{dark_blue}{\textbf{Silent Signals Dataset}} & Novel dataset of coded dog whistle examples created by applying the Dog Whistle Disambiguation task on the Potential Instance Dataset. & 13,220 & 3,330 \\
        \hline
        
    \end{tabular}
    \caption{Overview of the datasets used across experiments.}
    \label{table:data}
\end{table*}

\begin{figure}[t!]
    \centering
    \includegraphics[width=\columnwidth]{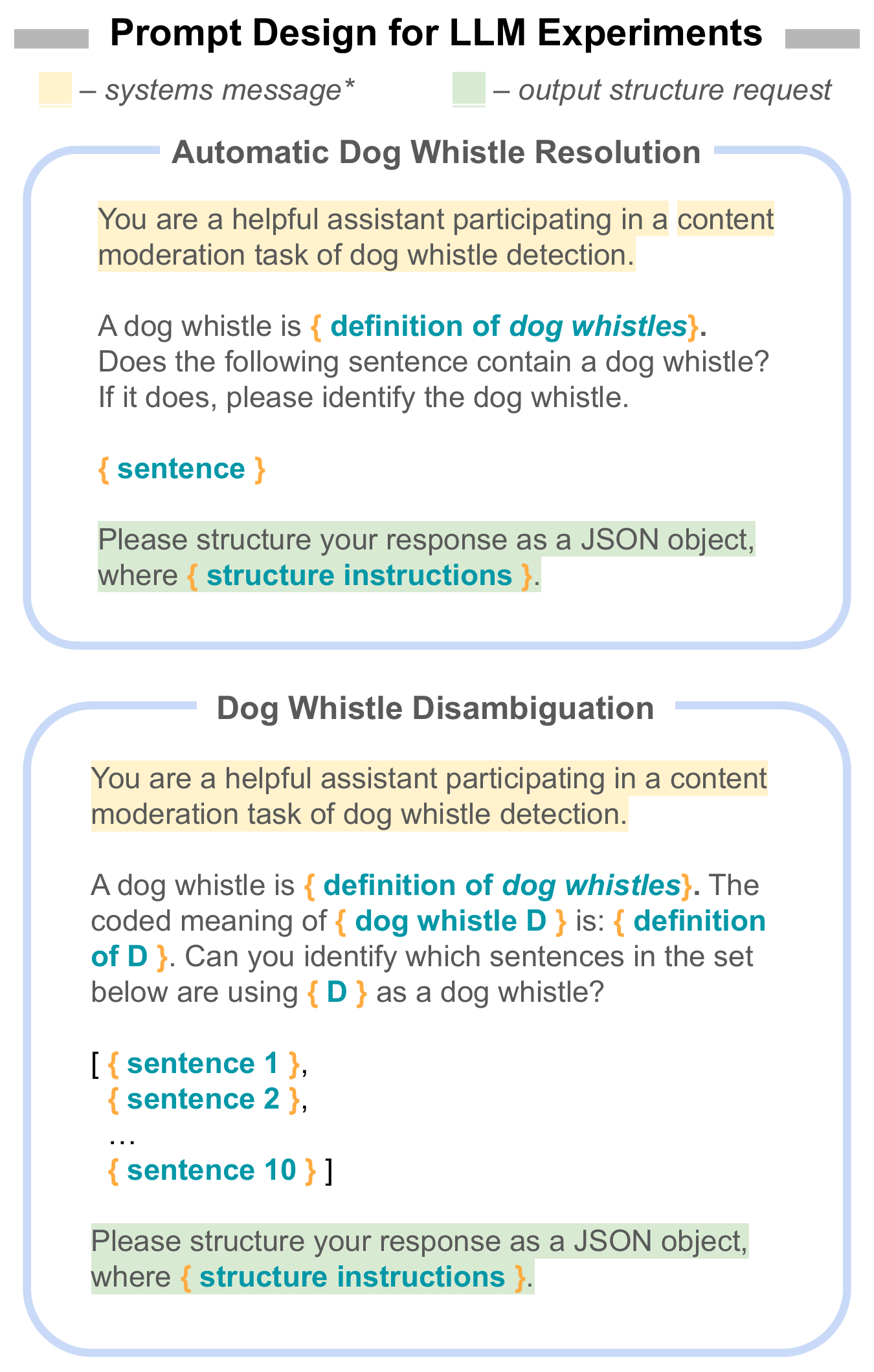}
    \caption{Visual representation of the different prompt structures used in \textit{Automatic Dog Whistle Resolution} (Section \ref{sec:baseline}) and \textit{Word-Sense Disambiguation} (Section \ref{subsec:wsd}) experiments. }
    \label{fig:prompt_vis_main}
\end{figure}

\section{Methods}
\subsection{Initial Data Collection}
To explore dog whistle disambiguation in both formal and informal settings, we pull public data from both Reddit and United States Congressional records. We collected Reddit comments from 2005-2022 in 45 controversial subreddits via the PRAW API and Pushshift archives \cite{baumgartner2020pushshift}. In addition to the Stanford Congressional Records dataset \cite{congress}, we use the \texttt{@unitedstatesproject} parser \cite{usproject} to compile congressional speeches from January 1900 to September 2023. For more details on data collection see Appendix \ref{appendix:data_collect}. 

\subsection{The \textit{Potential Instance} Dataset}
The Allen AI Glossary of Dogwhistles \cite{Mendelsohn2023FromDT} provides a list of 340 dog whistles with surface forms and examples to seed our keyword search for dog whistles iin the data we collected from Reddit and Congressional Records. We check for the presence of over 1000 surface forms to identify potential ``instances" of dog while use. Congressional entries which contain a keyword match are reduced to three sentence long excerpts. When a match is found in Reddit content, the entire submission or comment is retained. Each piece of content is annotated with the first key word found, so there may be more than one dog whistle present in one piece of text.

The resulting \textbf{Potential Instance} dataset spans approximately 6 million instances from Reddit comments, 1.1 million instances from Congressional records, and 327 distinct dog whistle. Entries in this dataset may be using the matched dog whistle phrase innocuously or with a coded meaning. At this step in the process, there is no way to separate the two cases.

\subsection{Synthetic Datasets for Evaluation}
\label{synthetic_detection}

We build two \textit{synthetic} datasets that are manually annotated for evaluation. The first, \textbf{Synthetic-Detection} contains 50 positive examples of single-word dog whistle terms from \citet{Mendelsohn2023FromDT}'s glossary, and 50 negative examples from Reddit and Congressional content, half of which contain an innocuous use of a dog whistle keyword, and the other half contain no keyword.

The second dataset, \textbf{Synthetic-Disambiguation}, contains 124 examples from Reddit and Congressional records which were manually annotated for evaluation. The dataset includes 13 distinct dog whistles, each with a corresponding set of 9-10 examples of this word used in discourse (with the exception of “jogger” which was added later with only 4 instances). These sets contain both coded and non-coded examples. This data was uniquely structured for the contrastive word-sense disambiguation task, where the model is provided a dog whistle, the definition of its coded meaning, and a set of ten sentences that contain that word or term. It is then prompted to separate sentences that use that word with the coded meaning from those that don't. A breakdown of the datasets used in this study can be found in Table \ref{table:data}.

All data was manually annotated by two members of our team with experience in this domain. The annotation was done individually, and only samples with agreement from both annotators were selected. Less than 5 samples were discarded due to disagreement.

\begin{table*}[!th]
    \centering
    \small
    \setlength{\extrarowheight}{2pt}
    \begin{tabular}{||P{0.8in} c | c | c c c c | c c c c||}
    \hline
         &  &  & \multicolumn{4}{c|}{\textbf{Zero-shot}} & \multicolumn{4}{c||}{\textbf{Few-shot}} \\
         &  & Human & GPT-3.5 & GPT-4 & Mixtral & Gemini &GPT-3.5 & GPT-4 & Mixtral & Gemini  \\[2pt] 
         \hline\hline
        \multirow{2}{*}{\parbox{0.9in} {\centering \vspace{3pt} \textbf{Presence}\\ \textit{\textcolor{dark_blue}{"is a dog whistle present?" }}}} & Acc & 66.8 & 80.0  & \textbf{85.0} & 68.0 & 81.0 & 76.0 & 86.6 & 81.0 & \textbf{86.7} \\[5pt]
        & F1 & 64.8 & 83.1 & \textbf{85.7} & 61.9 & 80.0 & 76.0 & 87.4 & 80.0 & \textbf{88.3} \\[5pt] 
        \hline
        \multirow{2}{*}{\parbox{0.9in} {\centering \vspace{3pt} \textbf{Identification} \\ \textit{\textcolor{dark_blue}{"identify the dog whistle."}}}} & Acc & 49.0 & 58.0 & 59.8 & 59.0 & \textbf{69.7} & 65.7 & 71.1 & 69.0 & \textbf{75.5}\\[5pt] 
        & F1 & 33.6 & 56.3 & 48.0 & 45.3 & \textbf{61.5} & 61.4 & 68.2 & 62.7 & \textbf{76.0}\\[5pt]  
        \hline
        \multirow{2}{*}{\parbox{0.9in} {\centering \vspace{3pt} \textbf{Definition} \\ \textit{\textcolor{dark_blue}{"define the dog whistle"}}}} & Acc & 47.3 & 52.0 & 54.6 & 58.0 & \textbf{66.7} & 60.6 & 67.0 & 67.0 & \textbf{73.5}\\ [5pt]
        & F1 & 29.7 & 46.7 & 37.1 & 43.2 & \textbf{56.0} & 53.0 & 61.9 & 59.3 & \textbf{73.5}\\[5pt] 
        \hline

    \end{tabular}
    \caption{Metric scores on the \textit{Automatic Dog Whistle Detection} task which surveys LLM and human ability to detect and define dog whistles in context. When presented with a sentence these experiments test the ability of a model/user to determine if the sentence contains a dog whistle and if so, correctly identify and define it. Predictions across all models have a statistical significance of $p<0.01$ by chi-squared test, and human predictions have statistical significant of $p<=0.037$. }
    \label{table:baseline}
\end{table*}

\section{LLM Experiments}
\label{sec:experiments}

\subsection{Automatic Dog Whistle Detection}
\label{sec:baseline}

If LLMs can reliably detect and explain political dog whistles with minimal annotation or engineering efforts, this would have significant implications for online moderation and computational social science, as this subtle form of hate speech detection could be at least partially automated at scale \citep{ziems2024can}. If LLMs match or outperform non-expert crowdworkers, then annotation resources could be shifted towards more efficient co-annotation between LLMs and experts \citep{li2023coannotating}. To test this, we evaluate four different models on the Synthetic-Detection dataset in a zero or few-shot manner.
Each model was provided with the definition of a political dog whistle and a candidate sentence, and was expected to predict whether the sentence included a dog whistle. If present, the model should then identify the span of text that contained the dog whistle and correctly define it.

Candidate models include GPT-3.5 \cite{brown2020language}, GPT-4 \cite{achiam2023gpt}, Gemini \cite{team2023gemini}, and Mixtral \cite{jiang2024mixtral}, as these have demonstrated strong performance on content moderation tasks \cite{jiang2024mixtral, buscemi2024chatgpt}. When prompt engineering on GPT-3.5,  we considered 5 different construct definitions and 3 additional phrasings of the prompt. We observed wide variation in performance, as in \citet{Mendelsohn2023FromDT}, but found that the Wikipedia definition of dog whistle and the following prompt was optimal: ``\textit{Does this sentence contain a dog whistle? If so, please identify it}''. Visualizations of prompt structure can be seen in Figure \ref{fig:prompt_vis_main}. For additional prompt engineering details, see Appendix \ref{appendix:prompts_auto}. 

\subsection{Human Baseline for Dog Whistle Detection}
\label{sec:human_eval} 
To ground the performance of LLMs on Automatic Dog Whistle Detection, we design a user study to establish a human baseline performance on this task. The aim is to gauge if LLMs could detect, identify, and define dog whistles better than a general populace. Leveraging the 100 test cases in the Synthetic-Detection dataset (Section \ref{synthetic_detection}), 62 Amazon Mechanical Turk workers were recruited to complete 720 unique annotations. Each annotation includes a classification on the presence of a dog whistle, and if applicable, identification of the dog whistle and its definition. Workers were expected to select dog whistle terms and definitions in isolation, from a set of 7 options (one of which was \textit{"I am not sure / not present in options"}) \footnote{This study was approved by the Georgia Institute of Technology Institutional Review Board (IRB)}. Workers were paid \$15/h. We vetted participants by inspecting their performance on non-coded negative examples. As half of the negative examples contained general speech, poor performance on these samples was deemed unlikely and indicative of poor quality annotation. This study included 28 individuals who identified as Male, 33 Female, and 2 Transgender, ages ranging from 18-22 to 50$+$. Participants in this study also declared diverse political views. For more information on annotators that contributed to this study, please see Appendix \ref{anno_demographics}.

\begin{figure*}[t!h]
    \centering
    \includegraphics[width=\textwidth]{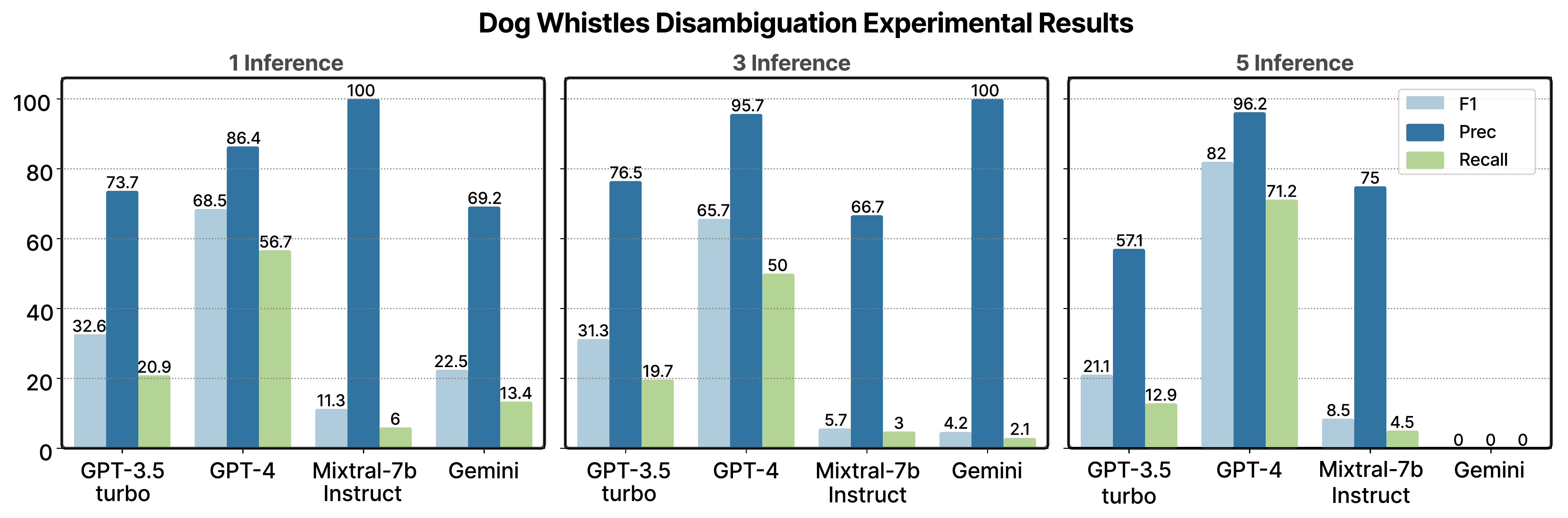}
    \caption{Results of \textit{Dog Whistle Disambiguation} task using the simulated ensemble across $N=1,3,5$ inferences. In an attempt to compensate for output volatility, for each N-inferences experiment, predictions are only considered if they remained consistent across all $N$ runs. Precision-1 and Recall-1 scores pertain to the positive class of coded dog whistle instances.}
    \label{fig:disambig_results}
\end{figure*}

\subsection{Dog Whistle Disambiguation}\label{subsec:wsd}
As preliminary investigation showed that LLM performance on Automatic Dog Whistle Detection left much to be desired (see Section \ref{sec:baseline}), we explore using these architectures for word-sense disambiguation. We leverage the Synthetic-Disambiguation dataset to evaluate LLMs' capacity to distinguish contexts in which a keyword appears with a harmful coded meaning from those in which the keyword appears innocuously. The prompt includes the Wikipedia definition of a dog whistle, the dog whistle keyword, and the word's coded meaning. The model performs classification for each of 10 example instances that contain the keyword, providing for each a label and an explanation for its decision.  

In an effort to improve the precision scores on the coded dog whistle instances, we simulate an ensemble-like approach where the model is prompted with the same task N consecutive times (as distinct chat completions) \footnote{Experiments were run with lower temperature values to prevent inference variation. However, decreasing temperature resulted in decreases in Precision of up to 10 points.}. Only predictions that have remained consistent over N inferences are kept, the others are discarded. We evaluate word-sense disambiguation of dog whistles over $N = 1, 3, 5$ consecutive inferences, as shown in Figure \ref{fig:disambig_results}. Specific details of prompt structure can be seen in Figure \ref{fig:prompt_vis_main}.

\section{Results}
\label{subsec:results}

Performance metrics from the \textit{Automatic Dog Whistle Detection} experiments (Section \ref{sec:baseline}) show that GPT-4 performed best on Dog Whistle Presence prediction in the zero-shot setting, and Gemini performed best on all other categories. However, no architecture produced remarkably high metrics on the Dog Whistle Definition task, for which the highest F1-score achieved with \textbf{Gemini} is \textbf{73.5}. For each model, there is a notable drop in performance as the complexity of the task increases from predicting the presence of a dog whistle, to identifying the dog whistle, and finally, defining it. For many examples, the model may correctly predict that a dog whistle is present, but incorrectly identify other provocative, but non-coded, language to be the dog whistle. Similarly, the model may correctly predict the presence of a dog whistle and correctly identify it in the text, but be unable to define it. This trend is also observed with the human baseline, which decreases with task complexity. Additionally, not that each LLM evaluated in this study out-performed the human baseline.

These initial investigations demonstrated that LLMs are unable to reliably identify and explain dog whistles.
Since these tasks are not solved, there remains a present need for larger training datasets with more numerous and varied examples of dog whistles. As described in Section \ref{subsec:wsd}, we explore applying LLMs to the task of word-sense disambiguation via prompting. The hypothesis is that providing the model with a set of examples would enable it to comparatively evaluate text and better disambiguate the coded instances from the non-coded. 

Although Gemini demonstrates superior performance on \textit{Dog Whistle Resolution}, GPT-4 achieves highest metric scores across all word-sense disambiguation experiments, especially when consistency in prediction for $N = $ 3 or 5 consecutive inferences is required. Whereas GPT-3.5 and GPT-4 respond well to this prompt structure, Gemini and Mixtral do not. Gemini's performance drastically decreases as the number of inferences increases, which is indicative of the architecture suffering from greater inference variation than other models in the study. Both Gemini and Mixtral are more reluctant to generate output in reference to potentially harmful content. With Gemini, the API explicitly blocked model output with code "block reason: other"\footnote{Outputs from Gemini were still blocked with this code after adjusting model safety settings to block none of the harassment and harm  categories.}. Mixtral would generate a response that expressed its inability to address the task. Examples that contained words such as "terrorists" (Gemini), "groomers" (Gemini), and "fatherless" (Mixtal) were common sensitivities. 

Most notably, increasing the number of consecutive inferences $N$ in the simulated ensemble approach for \textit{Dog Whistle Disambiguation} produced a precision score on coded dog whistle examples of \textbf{96.2} with \textbf{GPT-4} (as seen in Figure \ref{fig:disambig_results}). Although optimizing the precision score comes at the expense of recall, these experiments demonstrated that GPT-4 can be used to create a dataset of high confidence examples of coded dog whistle use. In Section \ref{sec:silent_sig}, we use this Dog Whistle Disambiguation method to create the Silent Signals dataset. 

\begin{figure}[t!]
    \centering
    \includegraphics[width=\columnwidth]{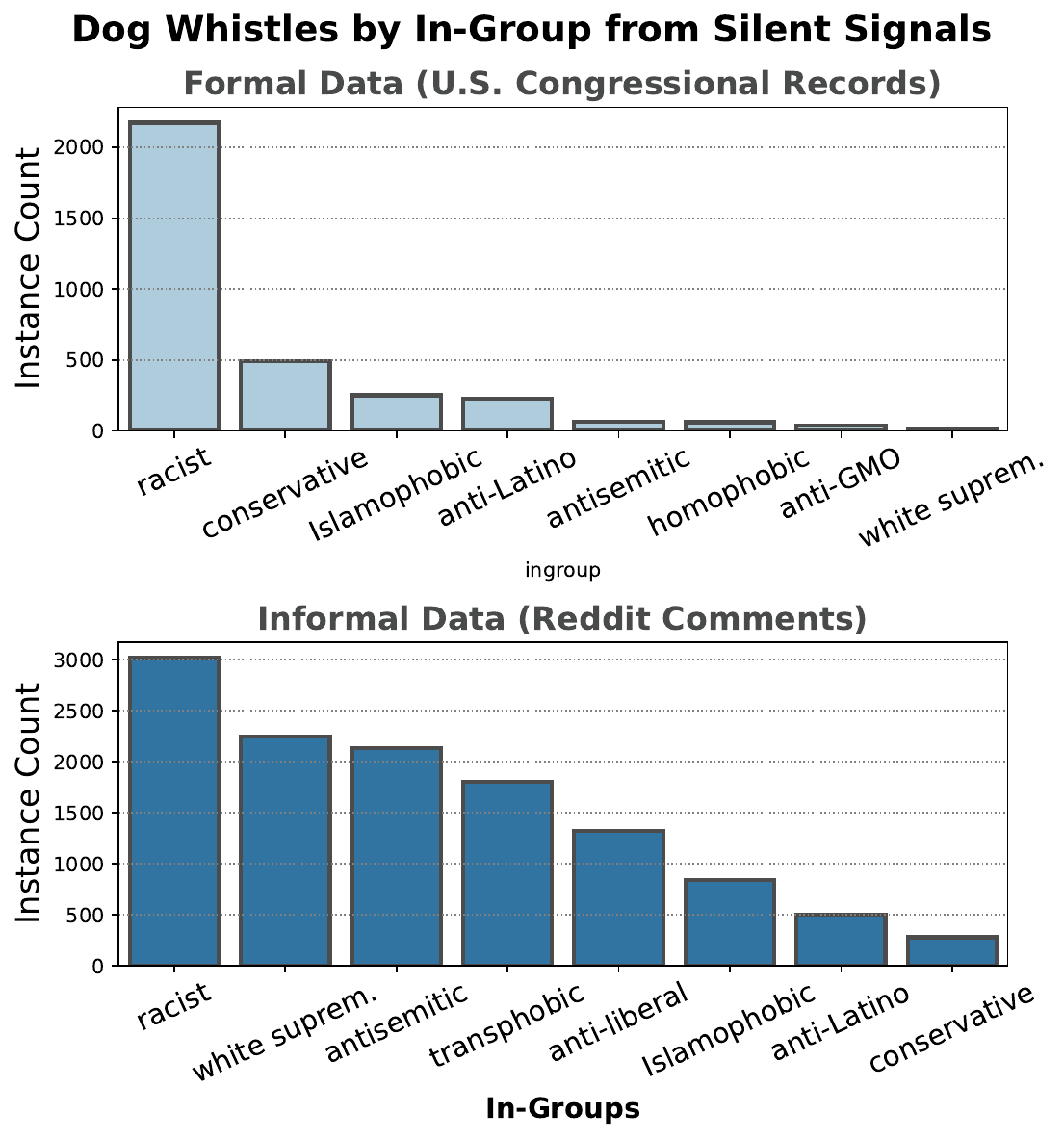}
    \caption{The distributions of dog whistles over in-groups for informal and formal communication in the Silent Signals dataset.}
    \label{fig:ss_ingroups}
\end{figure}

\begin{figure}[t!h]
    \centering
    \includegraphics[width=\columnwidth]{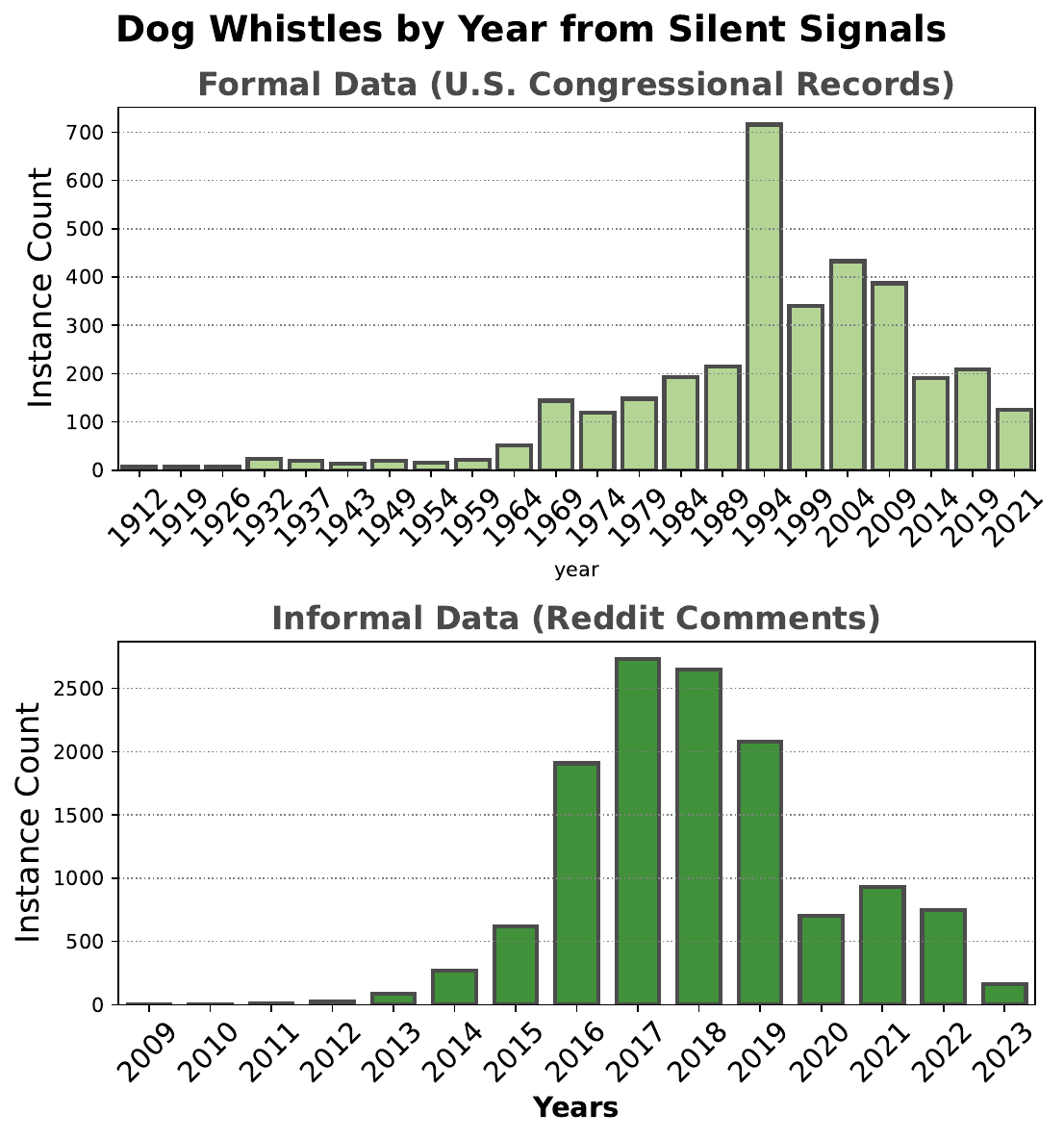}
    \caption{The distributions of dog whistles over time for informal and formal communication in the Silent Signals dataset.}
    \label{fig:ss_years}
\vspace*{-8pt}
\end{figure}

\section{Silent Signals Dataset}
\label{sec:silent_sig}
\citet{Mendelsohn2023FromDT}'s Dog Whistle Glossary documented a diverse collection of dog whistles across informal and formal communication. However, this resource alone does not address the challenges of conducting computational analysis of dog whistle use. Evaluating data based on key-word matches in text does not consider that many of those matches may not be coded uses of the dog whistle. To study the churn of dog whistles over time, their permeation through online communities and political parties, and their proliferation as vehicles for discriminatory speech, there must exist a means of disambiguation. 

Leveraging the word-sense disambiguation methodology presented in Section \ref{subsec:wsd} over 100,000 instances sampled randomly from the Potential Instance dataset, we create the Silent Signals dataset of high confidence coded dog whistle examples. We utilize the ensemble approach over 3 inferences with GPT-4. Information on dog whistles which were sampled at lower rates from the Potential Instance dataset can be found in Appendix \ref{appendix:keyword}. Each example in the Silent Signals dataset is annotated with the dog whistle present, dog whistle definition, type (formal or informal), in-group, and date. Congressional examples are also annotated with the chamber, speaker, and party, while Reddit instances are annotated with the subreddit. The dataset contains \textbf{\SSsize{}} instances across \textbf{298} dog whistles and \textbf{706} surface forms. Of these 79.8\% are informal instances from Reddit and 20.2\% are formal instances from Congressional speeches. The earliest dog whistle instance in the dataset dates to June 1, 1900 and the most recent to September 7, 2023.

\subsection{Validation} 
\label{subsec:valid}

In addition to our initial experiments which found a precision on coded dog whistle examples of 95.7\%, we manually evaluate a sample of 400 coded dog whistle examples in the Silent Signals dataset. This vetting procedure found a precision of 85.3\% amongst the positively predicted instances. However, for a number of these false positives, the word was in fact used as a dog whistle, but the coded meaning did not align with the definition provided in the Allen AI Glossary. For example, the glossary defines "terrorist" as an Islamophobic dog whistle with the coded implication that \textit{Muslim people on a whole are a threat}. In many instances captured in the Silent Signals dataset, however, "terrorist" is used not as an Islamophobic dog whistle but an anti-Liberal dog whistle. For example: "But they really turned splinter into a gay transpecies hedonist? The terrorists have truly won."\footnote{This post was shared in reference to the perceived queerness of the character Splinter in the 2023 movie Teenage Mutant Ninja Turtles: Mutant Mayhem.} In this instance, "terrorists" are implied to be liberals who support LGBTQ+ Rights. Taking into account these examples that do not fit the Allen AI definition but show signs of being novel dog whistle use, the accuracy over the vetted sample becomes 89.4\%. 

\begin{figure*}[th!]
    \centering
    \includegraphics[width=.95\linewidth]{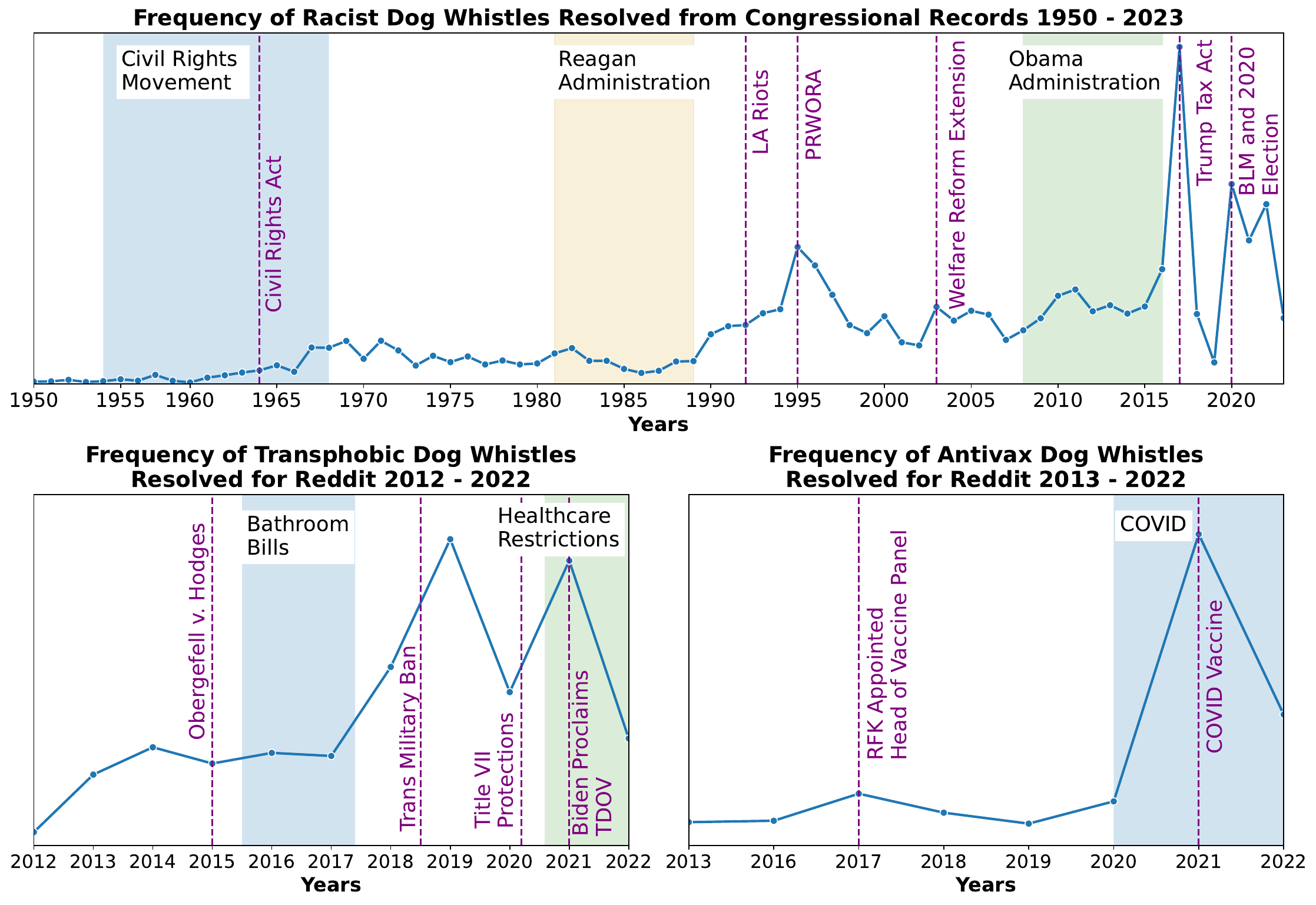}
    \caption{We investigate the use of Racist, Transphobic, and Anti-Vax dog whistles captured by the Silent Signals dataset over time. The graphs in this figure aree annotated with dates of pivotal political and cultural events in the United States.}
    \label{fig:ss_analysis}
\end{figure*}

\subsection{Analysis \& Characteristics}
The distribution of dog whistles in the Silent Signals dataset is visualized over in-group categories in Figure \ref{fig:ss_ingroups}, and over time in Figure \ref{fig:ss_years}. The sharp increase in dog whistles extracted from U.S. Congressional Records after 1960 aligns with the understanding that dog whistle use in politics gained popularity following the Jim Crow era \cite{mendelberg2001race, lasch2016sanctuary}. Furthermore, the disproportionately large amount of racist dog whistle detected in U.S. Congressional Records reflects political science research on historical use of dog whistles. Namely, that dog whistles were used to manipulate voter's racial animus after overt racism was not acceptable \cite{haney2014dog}. 

To demonstrate the utility of the Silent Signals dataset for political science research, we analyze the use of dog whistles over time by ingroup. Specifically, we graph and annotate the use of racist dog whistles in congressional records, as well as Transphobic and Anti-vax dog whistles on Reddit. As shown in Figure \ref{fig:ss_analysis}, the trends in dog whistles resolved per year demonstrate remarkable alignment with pivotal cultural and political events. The use of racial dog whistles in Congress increase from the beginning of the Civil Rights Movement, with significant peaks in years when Congress is discussing significant welfare and tax reforms such as the Personal Responsibility and Work Opportunity Reconciliation Act of 1996 (PRWORA). We also observe more recent spikes surrounding the 2016 Election, the Trump Tax Act, and the Black Lives Matter movement. Interestingly, we see that until the 1990s dog whistles do not have as large a presence in congressional speeches as might be anticipated, highlighting that political dog whistles were often used in more "public facing" venues such as speeches and campaign ads as opposed to on the Congress floor. With respect to Transphobic dog whistles, we see alignments in usage with significant points in the transgender and LGBTQ+ rights movements including \textit{Obergefell v. Hodge} (the Supreme Court Decision that required states to licence same-sex marriages), the passing of \textit{Bathroom Bills} (state legislation that denies access to public restrooms by gender identity), and the enactment of the \textit{Transgender Military Ban} during Donald Trump's presidency. The analysis of Anti-vax dog whistle usage on Reddit is, unsurprisingly, centered on COVID with a peak in 2021 following the availability of the COVID vaccine. The small exception to this is, interestingly, a brief increase in usage in 2017 surrounding the potential appointment by Donald Trump of known Anti-vaxxer Robert F. Kennedy to head a vaccine panel.

\section{Discussion}

\subsection{Evolution of Coded Meanings}
The validation of the Silent Signals Dataset enabled a salient observation of dog whistle use as it appears in Silent Signals. As discussed above, there were multiple cases in which a dog whistle was used with a covert meaning different from the definition present in the Allen AI glossary. Though this phenomenon was not frequent, it was far more common in colloquial instances than formal ones. This highlights the ways in which the study of neology is vital to the understanding of dog whistles given the rapid pace of linguistic change in online communities.

\subsection{Referential Speech}
\label{sec:ref_speech}
\vspace*{-4pt}

\begin{figure}[th!]
    \centering
    \includegraphics[width=\columnwidth]{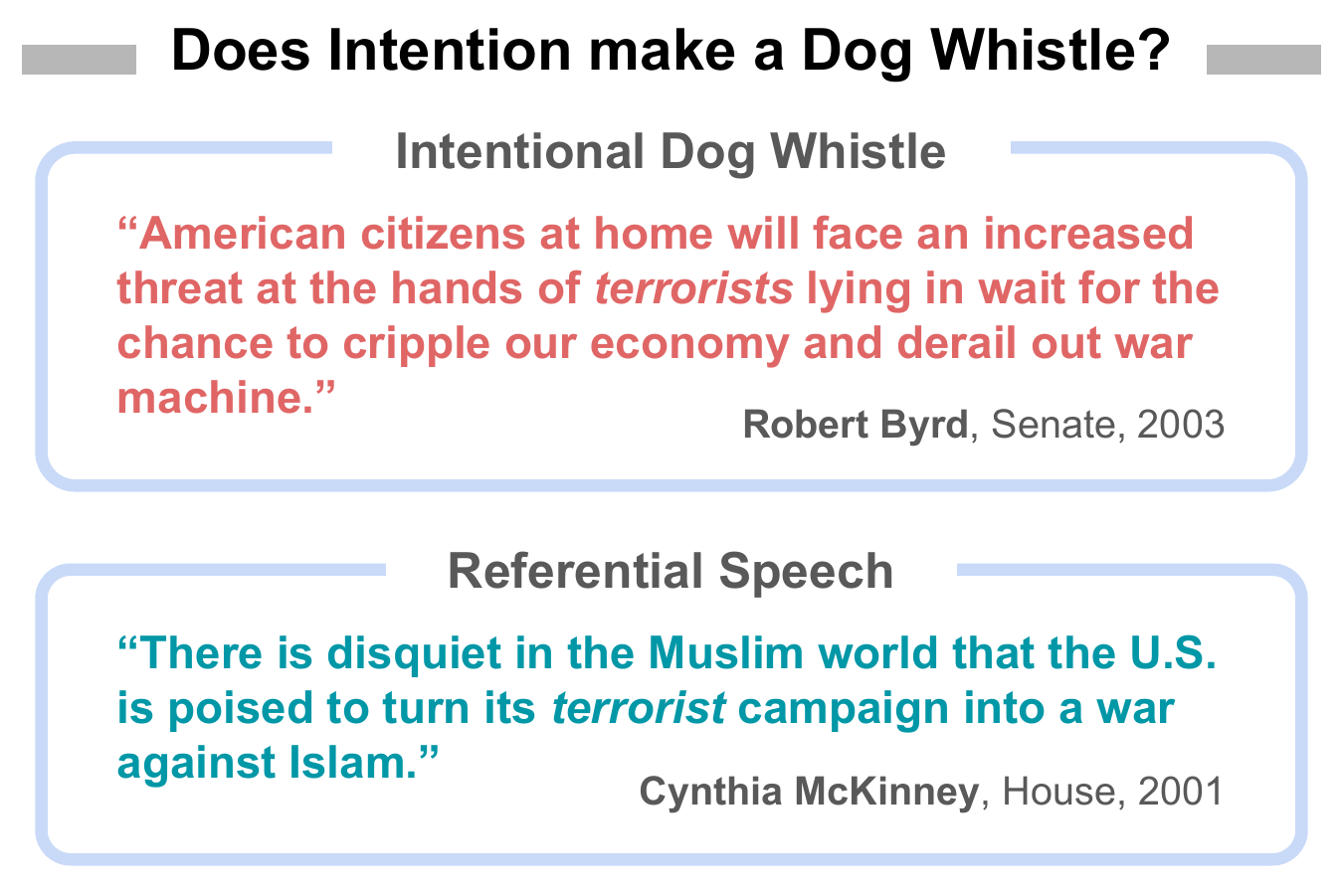}
    \caption{This figure demonstrates the difference between an intentional dog whistle and referential speech, using Congressional data in Silent-Signals.}
    \label{fig:ref_speech_exp}
\end{figure}

Among the 10.6\% of false positives identified during dataset validation, the vast majority were examples of referential speech or unintentional dog whistle use. These are cases where the term was used with the coded meaning, but without the explicit goal of discriminating against the target group. Figure \ref{fig:ref_speech_exp} demonstrates the difference with examples from the Silent Signals dataset. \textit{Terrorist} in both cases references the notion that \textit{Muslim people as a whole are a threat}. Where Senator Robert Byrd is using this word with this meaning intentionally, Representative Cynthia McKinney is attempting to draw attention to the War on Terror in US politics becoming increasing Islamophoic. There is not a strong consensus on how much intent makes a dog whistle. \citet{Saul2018} and \citet{quaranto2022dog} acknowledge unintentional dog whistles as a category of this coded speech, where as \citet{wetts2019called} and \citet{haney2014dog} regard intent as an integral component.

\subsection{Applications}
The Silent Signals dataset provides numerous opportunities for further study. It can be used to track dog whistle use over time, model the overlap between dog whistle use in formal and informal contexts, and investigate patterns of language used in various communities, virtual and other wise. From a Political Science perspective, it provides opportunity for analysis of dog whistle use along partisan and speaker-based axes. It can be used to explore how dog whistle use corresponds with social and political movements in the United States. From a Machine Learning perspective, Silent Signals can be used to training and/or finetuning could be performed for tasks ranging from hate speech detection to emergent dog whistle identification.

\section{Conclusion}
 Dog whistles are used to promote discrimination in both formal and informal environments. The use of this coded language allows speakers to maintain plausible deniability and bypass hate speech detection systems when used online. 
 This work presents the largest, to date, survey of LLM capabilities with respect to the automatic resolution of dog whistles.  
 Experimental results demonstrate that LLMs remain unreliable in the dog whistle resolution task. A hindrance to research in this space has been the unavailability of large datasets of coded dog whistles examples. We show that despite the overall inconsistencies of LLMs on the automatic dog whistle resolution task, with the proper methodology, they are adept at disambiguating coded dog whistles from standard language. We leverage this capability to create the  Silent Signals dataset which contains \SSsize{} dog whistle examples and 298 distint dog whistles. We believe that this resource will be integral to the continued study of dog whistles with applications in content moderation, computational social science, and political science, on tasks such as analysis of trends in dog whistle use, dog whistle resolution, hate speech detection, and identification of emergent dog whistles.

\section{Limitations}
Language naturally evolves as it permeates through communities. In the case of dog whistles, this can result in a broadening or changing of target groups, terms, or coded meanings. While the Allen AI glossary is a foundational work without which this research would not be possible, it likely does not encompass all dog whistles, use cases, and definitions. As such, though the Allen AI glossary and the Silent Signals dataset both provide helpful tools for the continued research of dog whistles, the rapidly evolving nature of coded language can render these resources outdated and incomplete. Further, there is the question of whether dog whistles are always used intentionally or simply perpetuate harmful tropes the speaker may be unaware of. \citet{susihChillEmbargo} explore this idea in the context of the dog whistle "bankers":  \textit{"Were the Populists’ attacks on greedy bankers—some of which used terms like Shylock or invoked the Rothschilds—meant to focus anger and hatred on the Jews, or was the association so sublimated that the Populists didn’t even realize they were blowing a dogwhistle?"} . Though we include such use cases in the Silent Signals dataset, it remains unclear to what extent intentionality defines the dog whistle.

Additionally, please note that the dog whistles documented in the Allen AI glossary are not evenly distributed over all in-groups or between formal and informal speech. For example, racist dog whistles, which are most common in both the informal and formal portion of Silent Signals are also the most common in-group in the glossary. White supremacist dog whistles, on the other hand, are resolved at a very low rate from the congressional data have only 5 formal entries present in the glossary. As a result, the distribution of dog whistles over in-group in Silent Signals will reflect these biases and is not necessary reflective of dog whistle usage as a whole. Specific breakdowns of the Allen AI glossary by ingroup and formal/informal sphere can be seen in Appendix \ref{appendix:allen_ai}.

From a computational perspective, our method achieved high precision on the Dog Whistle Disambiguation Task. However, optimizing on precision comes at the expense of recall. Improving the efficiency of word-sense disambiguation with LLMs remains an open problem. Additionally, using GPT-4 in the creation of Silent Signals subjects it any biases in the model. We recognize that we may have resolved specific types of dog whistles more frequently than others. 

With respect to establishing a human baseline, a natural extension of this work would be to investigate how LLM performance would compare to domain expert prediction, and how annotator attributes correlate with understanding of political coded language. Observations from our concise user study suggest that human performance on this task may vary with context behind the coded language and the community the individual belongs to. Due to resource constraints, we were not able to collect enough annotations from various communities to derive any statistically significant trends on these axes. We hope to address these pertinent investigations in future work.

Finally, although we collect over 7 million potential dog whistle instances, due to resource constraints, we only sample 100,000 instances for the creation of the Silent Signals dataset. Therefore, we release the Potential Dog Whistle Dataset to enable the open sourced expansion of the Silent Signals dataset.

\section*{Acknowledgements}  We are thankful to the anonymous reviewers, as well as members of SALT Lab for their helpful feedback on various stages of the draft. This work was partially supported by a grant from Meta. Caleb Ziems is supported by the NSF Graduate Research Fellowship under Grant No. DGE-2039655.

\label{sec:bibtex}

\bibliography{acl_latex}

\clearpage

\appendix

\section{Data Collection Details}
\label{appendix:data_collect}

\subsection{Reddit}
\label{appendix:reddit}
Subreddits included as a part of the Potential Instance and Silent Signals datasets.
\begin{table}[!th]
    \centering
    \small
    \setlength{\extrarowheight}{2pt}
    \begin{tabular}{c c}
        4chan & Antiwork  \\
        AsianMasculinity & aznidentity \\
        BlackPeopleTwitter & Braincels \\
        CBTS\_stream & ChapoTrapHouse \\
        Chodi & climateskeptics\\
        conservatives & consoom \\
        conspiracy & Coontown  \\
        CringeAnarchy & European \\
        FemaleDatingStrategy & frenworld \\
        GenderCritical & GenderCynical  \\
        GenZedong & GoldandBlack \\
        GreatAwakening & HermanCainAward \\
        incels & IncelsInAction  \\
        KotakuInAction & MensRights \\
        MGTOW & MillionDollarExtreme \\
        Mr\_Trump & NoFap \\
        NoNewNormal & Portugueses  \\
        prolife & Russia \\
        RussiaPolitics & SocialJusticeInAction  \\
        The\_Donald & TheRedPill \\
        TrueUnpopularOpinion & TruFemcels \\
        TumblrInAction & UncensoredNews  \\
        walkaway & WhitePeopleTwitter \\

    \end{tabular}
\end{table}

\subsection{Keyword Matching Considerations}
\label{appendix:keyword}
There were a select few dog whistles which occurred at incredibly high rates in the non-coded sense. Due to resource constraints, we did not want to expend large amounts of compute on dog whistles which where most commonly used innocuously. As such, a select few dog whistles were excluded or sampled at a lower rate for the creation of the Silent Signals dataset. In the Congressional dataset, the dog whistles "XX", "federal reserve", "based", and "single" were excluded due to their high rate of innocuous usage and the fact that initial surveys indicated no coded uses. In the Reddit dataset, the dog whistles "based" and "single" were down sampled based on the frequency of their non-coded use in the Potential Instance dataset. Importantly, even with this down sampling, the Silent Signals dataset still contains coded instances of both "based" and "single".

\section{User Study on Dog Whistle Detection}

The table below outlines the demographic attributes of participants in this study. Note that participants span many ages, genders, and political views. Due to the nature in which coded language like dog whistles is disseminated, it is likely that attributes such as political affiliation and preferred social media platform may correlate with literacy in specific kinds of coded speech. However, this user study did not collect annotation from enough individuals to male statistically significant findings in this area.

\begin{table}[!th]
    \setlength{\arrayrulewidth}{0.2mm}
    \centering
    \small
    \setlength{\extrarowheight}{3pt}
    \begin{tabular}{| c | c | c | c | c | c | c | c |}
        \hline
        \multicolumn{4}{|c|}{\textbf{\textcolor{dark_blue}{Annotator Demographics}}}\\
        \hline
        \multicolumn{2}{|c|}{\textbf{Gender}} & \multicolumn{2}{c|}{\textbf{Age}} \\
        \hline

        Woman & 33 & 18 - 22 & 1 \\
        Man & 28 & 22 - 30 & 22 \\
        Transgender & 2 & 30 - 40 & 16 \\
         & & 40 - 50 & 17 \\
         & & 50 + & 7 \\

        \hline
        \multicolumn{2}{|c|}{\textbf{Ethnicity}} & \multicolumn{2}{c|}{\textbf{Religion}} \\
        \hline

        White & 53 & Christianity & 48 \\
        Asian & 3 & Atheist/Agnostic & 13 \\
        Native American & 3 & Other & 2 \\
        Black & 2 & & \\
        Latino & 2 & & \\
        
        \hline
        \multicolumn{2}{|c|}{\textbf{Political Views}} & \multicolumn{2}{c|}{\textbf{Education}} \\
        \hline

        Liberal & 24 & HighSchool & 5 \\
        Conservative & 22 & Associates/Trade & 5 \\
        Moderate & 12 & Bachelors & 35 \\
         & & Masters & 18 \\
    
        \hline
        \multicolumn{4}{|c|}{\textbf{Preferred Social Media}} \\
        \hline

        \multicolumn{2}{|c|}{Instagram} & \multicolumn{2}{c|}{31} \\
        \multicolumn{2}{|c|}{Twitter} & \multicolumn{2}{c|}{8} \\
        \multicolumn{2}{|c|}{Reddit} & \multicolumn{2}{c|}{7} \\
        \multicolumn{2}{|c|}{TikTok} & \multicolumn{2}{c|}{3} \\
        \multicolumn{2}{|c|}{Other} & \multicolumn{2}{c|}{2} \\
        
        \hline
        
    \end{tabular}
    \caption{This table reports the demographic attributes of annotators that contributed to the human baseline on Dog Whistle Detection. See Section \ref{sec:human_eval}.}
    \label{anno_demographics}
\end{table}

\section{Prompting}
\label{appendix:prompts} 
\begin{figure}[t!h]
    \centering
    \includegraphics[width=\columnwidth]{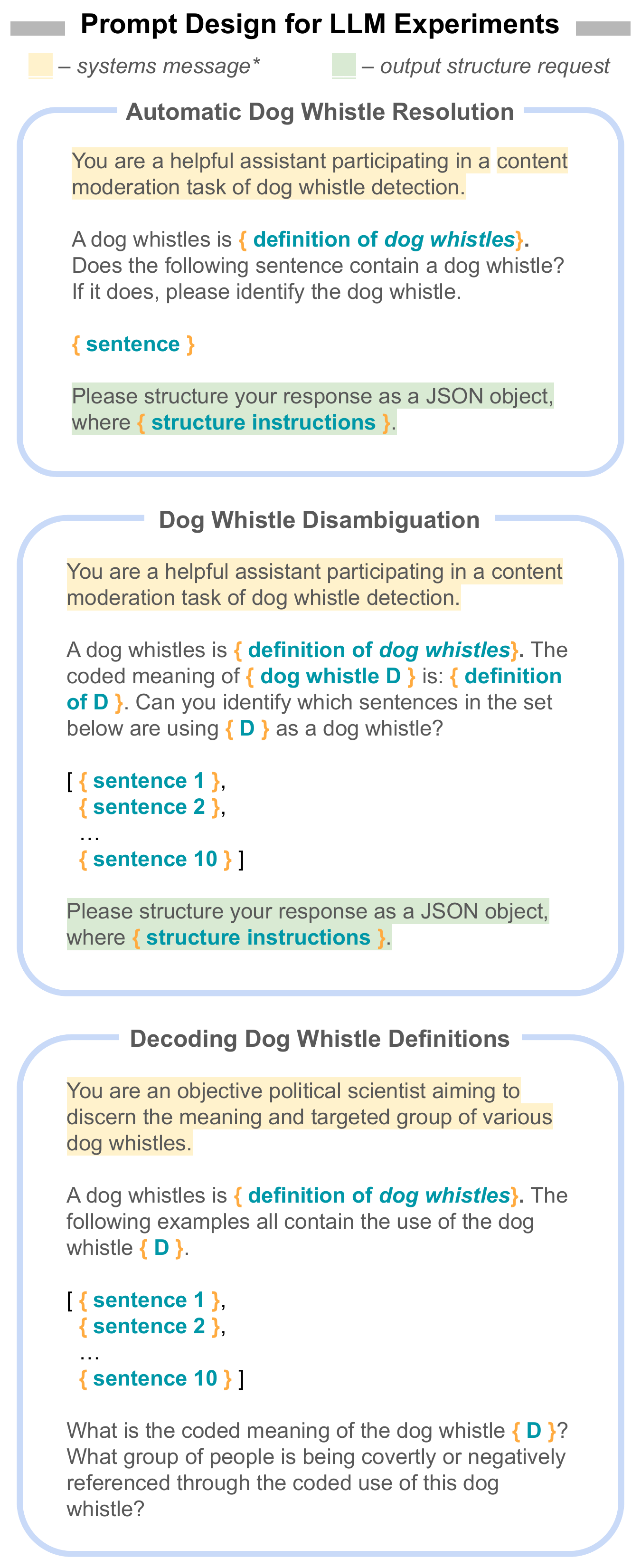}
    \caption{Visual representation of the different prompt structures used in \textit{Automatic Dog Whistle Resolution} (Section \ref{sec:baseline}),  \textit{Word-Sense Disambiguation} (Section \ref{subsec:wsd}), and \textit{Decoding Dog Whistle Definition} (Appendix \ref{appendix:def}) experiments. }
    \label{fig:prompt_vis}
\end{figure}

The structure of prompts for all LLM experiments can be found in Figure \ref{fig:prompt_vis}.

\subsection{Automatic Dog Whistle Detection}
\label{appendix:prompts_auto}
While in the prompt engineering stages of our work, we ran a number of experiments on GPT-3.5 to determine which combination of dog whistle definition and prompting question would produce the best results. Specifically, we tested 5 dog whistle definitions and 3 questions. Results of these experiments can be seen in Table \ref{table:prompt_exp} and Table \ref{table:question_exp} respectively. Due to their high rate of dog whistle resolution the definition \textit{"A dogwhistle is the use of coded or suggestive language in political messaging to garner support from a particular group without provoking opposition."} and the prompt \textit{"Does the following sentence contain a dog whistle? If it does, please identify the dog whistle."} were selected and used throughout our work.

Our experiments with GPT-3.5 informed our selection of the optimal definition of the term "dog whistle". In further stages of this work, however, we adapted prompt structures to the other models used in the study.

\begin{table*}[t!h]
    \centering
    \small
    \setlength{\extrarowheight}{2pt}
   \begin{tabular}{||p{0.37\linewidth} |p{.08\linewidth} p{0.1\linewidth} p{0.12\linewidth} p{0.1\linewidth}||}
    \hline
        ~ & No Dog Whistle Detected & Incorrect Dog Whistle Identified & Correct Dog Whistle, Incorrect Definition & Correct Dog Whistle and Definition \\ \hline\hline
        \textbf{Does the following sentence contain a dog whistle?} & 20.0 & 24.5 & 28.2 & 26.4  \\
        \textbf{Does the following sentence contain a dog whistle? If it does, please identify the dog whistle.} &  \textbf{8.0} & \textbf{19.0} & \textbf{22.0} & \textbf{51.0} \\
        \textbf{Does the following sentence contain a dog whistle? If it does, please identify the dog whistle and describe what it secretly means.}&  7.1 & 20.2 & 23.23 & 49.5 \\
        \hline

    \end{tabular}
    \caption{Analysis of GPT-3.5 output across 3 prompting questions. Given it had the highest rate of dog whistle resolution, the second prompt was selected as the prompting question for the automatic dog whistle resolution task.}
    \label{table:question_exp}
\end{table*}

\section{LLM Behavioral Trends}
In the process of conducting the experiments described in Section \ref{sec:experiments}, the following behavioral trends were observed for the models evaluated. We provide this information as a guide for practitioners who may seek to conduct similar investigations:

\begin{enumerate}
    \item GPT struggled with performance when output structures were requested. Specifically, we saw our performance decrease ~3-5 points when output was requested to be formatted in JSON or list form.
    \item When asked to provide its reasoning, we witnessed a 5-10 point increase in performance across models.
    \item Certain models are more and less amenable to certain prompt structures. Specifically, Gemini and Mixtral struggled greatly with multi-example prompts where multiple instances were requested to be interacted with in a single run (for example in the word sense disambiguation task when multiple instances needed to be categorized).
    \item Gemini was only usable for this task after all user safety blocks had been disabled. Even with these blocks disabled, there were still a number of cases in which the model blocked output by throwing an error messsage.
    \item Mixtral was only cooperative once "This is a content moderation task" was included in the prompt.
\end{enumerate}

\begin{table*}[h!]
    \centering
    \small
    \setlength{\extrarowheight}{3pt}
   \begin{tabular}{||p{0.37\linewidth} |p{.08\linewidth} p{0.1\linewidth} p{0.12\linewidth} p{0.1\linewidth}||}
    \hline
        ~ & No Dog Whistle Detected & Incorrect Dog Whistle Identified & Correct Dog Whistle, Incorrect Definition & Correct Dog Whistle and Definition \\ \hline\hline
        \textbf{A dogwhistle is an expression that has different meanings to different audiences.} \cite{Albertson2014} & 7.8 & 29.7 & 23.4 & 39.1  \\
        \textbf{A dogwhistle is a word or phrase that means one thing to the public at large, but that carry an additional, implicit meaning only recognized by a specific subset of the audience.} \cite{Bhat2020} &  15.9 & 22.2 & 22.2 & 39.7 \\
        \textbf{A dogwhistle is a term that sends one message to an outgroup while at the same time sending a second (often taboo, controversial, or inflammatory) message to an ingroup.} \cite{Henderson2018}&  11.1 & 27.0 & 23.8 & 38.1 \\
        \textbf{A dogwhistle is a coded message communicated through words or phrases commonly understood by a particular group of people, but not by others.} \cite{mw:dw}& 17.5 & 25.4 & 22.2 & 34.9 \\
        \textbf{A dogwhistle is the use of coded or suggestive language in political messaging to garner support from a particular group without provoking opposition.} \cite{wiki:Dog_whistle_(politics)} & \textbf{6.5} & \textbf{25.8} & \textbf{25.8} & \textbf{41.9} \\
        \hline

    \end{tabular}
    \caption{Analysis of GPT-3.5 output across 5 dog whistle definitions. Given it had the lowest rate of detecting no dog whistles and the highest rate of correctly resolving dog whistles, the Wikipedia definition was selected as the definition used throughout the rest of our experiments. }
    \label{table:prompt_exp}
\end{table*}

\section{Glossary Analysis}\label{appendix:allen_ai}
\begin{figure*}[th!]
\includegraphics[width=\linewidth]{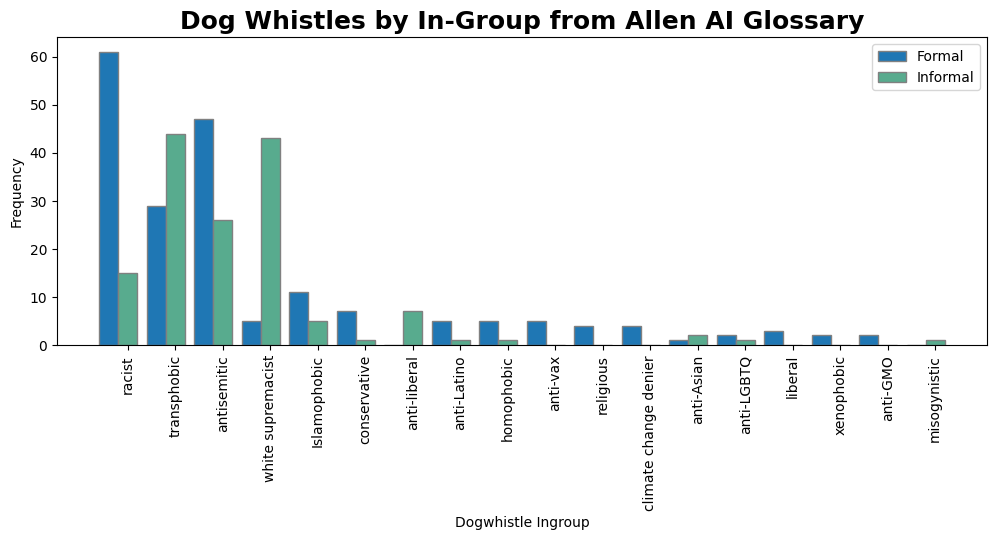}
    \caption{This graph shows the distribution of dog whistles in the Allen AI glossary by ingroup and sphere of use.}
    \label{fig:allen_ai}
\end{figure*}

The Allen AI Glossary of Dogwhistless contains 340 English-language dog whistles from informal and formal spheres. As the glossary is designed as a tool to document and define dog whistles, there is not an even split between informal and formal dog whistles present nor an even distribution amongst ingroups. In total there are 193 formal and 147 informal dog whistles documented in the Allen AI glossary. The distribution of both formal and informal dog whisltes by ingroup can be seen in Figure \ref{fig:allen_ai}. These uneven distributions are important to keep in mind when viewing the distribution of dog whistles by ingroup in the Potential Instance and Silent Signals dataset, as some ingroups may have higher rates of representation in the datasets because of their high representation in the Allen AI glossary.

\section{Further Dog Whistle Definition Experiments}
\label{appendix:def}
Following our initial survey of LLM performance on automatic dog whistle resolution, we explored means of improving the architectures' ability to decode hidden meanings of dog whistles. To do so we provide the model with additional context in the form of multiple coded examples of a specific dog whistle from the Synthetic-Disambiguation dataset. Specifically, the model is given a definition of what a dog whistle is, the dog whistle to be evaluated, and a set of 3 - 7 coded examples of the dog whistle in context and is asked to return the coded meaning of the dog whistle. For specific prompting details see Figure \ref{fig:prompt_vis}. As a point of comparison, we run a parallel experiment in which no example dog whistle instances are provided as a means to gauge the effect that additional context has on the LLMs' ability to accurately define dog whistles. This experiment is run on the Synthetic-Detection dataset and exclusively with GPT-4, as this model was most amenable to the multi-example setting.

Predictions shown in Table \ref{table:experiment2} were manually validated referencing definition and targeted group information provided in the Allen AI glossary. To allow for nuance, we evaluate each predicted definition on a scale of 0 to 2, where 0 is \textit{incorrect}, 1 is \textit{incomplete}, and 2 is \textit{correct}. Incomplete definitions of dog whistles or their targeted groups are characterized by mis-identification of the target group, incorrect implications of the term, or failure to underline the harmful nature of the coded speech. For example, an incomplete definition might say a dog whistle carries connotations that are "anti-political correctness, non-conformity, anti-establishment" as opposed to connotations of alt-right or white supremacist views. 

\begin{table}[!b]
    \centering
    \small
    \setlength{\extrarowheight}{2pt}
    \begin{tabular}{|l|c c|}
    \hline
        ~ & \% Fully & \% Correct \\ 
        ~ & Correct & (w/ Incomplete) \\ \hline
        \multicolumn{3}{|c|}{\textcolor{dark_blue}{\textbf{no context}}} \\ \hline
        Definition & 69.2 & 84.6  \\
        Targeted Group &  53.8 & 69.2 \\
        Definition and Group &  61.5 & 84.6 \\ \hline
        
        \multicolumn{3}{|c|}{\textcolor{dark_blue}{\textbf{in context}}} \\ \hline
        Definition & 69.2 & 92.3 \\
        Targeted Group & 53.8 & 69.2 \\
        Definition and Group & 69.2 & 92.3 \\ \hline

    \end{tabular}
    \caption{Ability of GPT-4 to accurately define dog whistles and their target group. No context experiments present only the dog whistle while in context experiments present the dog whistle along with 3-7 coded examples of its use. Partially correct responses may identify part but not all of the definition or target group or else fail to underline the hateful and harmful nature of the given dog whistle.}
    \label{table:experiment2}
\end{table}

The \textit{Decoding Dog Whistle Definitions} experiment was designed with the hypothesis that providing a model with multiple examples of a dog whistle's usage would improve its ability to resolve the definition. However, when counting only fully correct responses, there is very little difference between results when only a dog whistle was presented and results when we provided the dog whistle and 3-7 coded instances of its use. When including partially correct definitions, the addition of examples had greater impact on model output. Best results were found when prompting the model to identify both the definition and target group, while the model struggled most to identify only the targeted group of a given dog whistle.

\end{document}